\documentclass[letterpaper, 10 pt, conference]{ieeeconf}  

\IEEEoverridecommandlockouts                              

\usepackage{amsmath} 
\usepackage{amssymb}  
\usepackage{graphicx} 
\usepackage{multirow}
\usepackage{verbatim}
\usepackage{xcolor}
\usepackage{subfigure}

\usepackage[ruled,vlined]{algorithm2e}
\usepackage{algpseudocode}
\usepackage{comment}

\title{\LARGE \bf
Socially-Aware Opinion-Based Navigation with Oval Limit Cycles
}

\author{Giulia d'Addato$^{1,3}$, Placido Falqueto$^{2}$, Luigi Palopoli$^{2}$, Daniele Fontanelli$^{3}$ 
\thanks{}
\thanks{$^{1}$ Project co-funded by the European Union – Next Generation Eu - under the National Recovery and Resilience Plan (NRRP), Mission 4 Component 1 Investment 4.1 - Decree No. 118 (2/03/2023)  of Italian Ministry of University and Research - Concession Decree No. 2333 (22/12/2023) of the Italian Ministry of University and Research, Project code D93C23000470005, within the Italian National Program PhD Programme in Autonomous Systems (DAuSy).}
\thanks{$^{2}$ Department of Information Engineering and Computer
  Science, Universit\`a  di Trento, Italy.}
\thanks{$^{3}$ Department of Industrial Engineering, Universit\`a di
  Trento, Italy.} 
}

\newcommand{\FriW}[1]{\emph{FriWalk}}

\newcommand{\rev}[1]{\color{red}{{\bf #1}}\color{black}}

\begin{document}

\maketitle
\thispagestyle{empty}
\pagestyle{empty}

\begin{abstract}
When humans move in a shared space, they choose navigation strategies that preserve their mutual safety. At the same time, each human seeks to minimise the number of modifications to her/his path. In order to achieve this result, humans use unwritten rules and reach a consensus on their decisions about the motion direction by exchanging non-verbal messages. They then implement their choice in a mutually acceptable way. Socially-aware navigation denotes a research effort aimed at replicating this logic inside robots. Existing results focus either on how robots can participate in negotiations with humans, or on how they can move in a socially acceptable way.
We propose a holistic approach in which the two aspects are jointly considered. Specifically, we show that by combining opinion dynamics (to reach a consensus) with vortex fields (to generate socially acceptable trajectories), the result outperforms the application of the two techniques in isolation.

\end{abstract}


\section{INTRODUCTION}

An increasing number of applications requires mobile robots to move through human-populated areas. Significant examples include automated guided vehicles (AGVs) used for intra-factory logistics and warehouse management, mobile robots used for shipping and delivery in urban environments, and assistive robots used to support the mobility of disabled and elderly users.
The nature of these applications requires the robot to accomplish its task reliably and efficiently. At the same time, the safety of human bystanders must not be imperilled. Last but not least, the trajectories followed by the robot should be perceived as smooth and human-friendly as those followed by its human counterparts (this is even truer when the robot is carrying or guiding a human).
To generate this kind of trajectory on a robot, we need to take inspiration from what humans do. When humans move in a shared and crowded space, they exchange a short but intense stream of non-verbal signals to decide which direction each of them should take. Once each human makes up her/his mind on the direction to follow, s/he implements the choice by modifying the trajectory that s/he will follow in the next few seconds. This modification minimises the energy spent and the strain on the junctures \cite{lavalle2001nonholonomic}, but at the same time it respects the private space of the other humans \cite{marquardt2015proxemic}.
This complex set of communication protocols, social rules and reactive adjustments to our trajectories operates in strict coordination. They are so deeply embedded in our social behaviour that their use hardly touches the conscious level.
Socially-aware navigation, or simply social navigation, is a research area that seeks to approximate the logic of human behaviour summarised above on a robot. 

\noindent 
{\bf Related work.} An intense research activity on social navigation in the past few years has delivered different strategies aimed at ensuring a safe and efficient interaction between robots and humans~\cite{socialGroups,falqueto2024humanising}. Research in this area includes various approaches designed to address the complexities of navigating in dynamic environments shared with humans, with both proactive and reactive approaches being explored~\cite{socialNavig_survey,conflictsSocialNavig}. In short, proactive strategies attempt to predict and adapt to human behaviour, while reactive strategies respond to real-time environmental changes.

Reactive methods are essential when people in a crowd move depending on the surrounding situation, requiring rapid re-planning of the robot's trajectory. However, as the number of people increases, accurate and fast path planning becomes more difficult, increasing the computational time required to find the optimal path. More computationally efficient reactive methods are artificial potential fields~\cite{potentialFields,ovalBoldrer}, the velocity obstacle (VO)~\cite{velocity_obstacle}, and the social force model (SFM)~\cite{sfm_ferrer,helbing_social}. SFM describes pedestrian movement as influenced by social forces that guide acceleration towards desired velocities while maintaining interpersonal distances. Although effective for large groups, SFM struggles with individual-level interactions in open environments. Enhanced models, such as the Headed Social Force Model (HSFM)~\cite{hsfm_fontanelli}, incorporate pedestrian heading to better predict non-holonomic motion.

Proactive strategies, on the other hand, account for mutual interactions in crowded situations, enabling robots to cooperate with humans. These approaches involve predicting human behavior and proactively planning a collision-free path. Examples include proactive models based on SFM~\cite{proactiveSFM}, opinion dynamics~\cite{opinion_leonard}, and dynamic path planning techniques such as the Morphing algorithm~\cite{morphingAlg}.

Sampling-based motion planning methods, such as Rapidly-exploring Random Tree (RRT)~\cite{lavalle2006planning,silva_rrt} and Risk-RRT~\cite{risk_rrt_navigation}, are also commonly used for dynamic environments. However, these methods can struggle with the complexities of interactive behaviors in human-populated spaces. Finally, learning-based approaches, including reinforcement learning (RL)~\cite{rl_socialNav}, deep RL (DRL)~\cite{deepRL}, and inverse RL (IRL)~\cite{irl_socialNav,irl_baghi}, are increasingly being explored for crowd navigation. These methods are promising but often lack analytical tractability and can be computationally demanding in real-time scenarios.

\noindent
{\bf Paper Contribution.}
In the short report above, we have mentioned two important classes of work: proactive and reactive path planning. In the first class, the robot negotiates with humans the direction of motion that each should take, but the implementation of the decision is not guaranteed to be safe and does not meet the quality standards that the human requires. In the second, the decision is taken and the robot concentrates on implementing it through a human-friendly trajectory. The missing piece in this case is a mechanism that generates the ``strategic" decision. 

In this paper, we propose to merge the two phases into a unified holistic approach.
We use social dynamics to implement the negotiation phase~\cite{opinion_leonard}, \emph{and} potential fields with oval-shaped limit cycles to generate a socially-aware trajectory~\cite{ovalBoldrer}. The opinion dynamics module generates a variable that encodes the decision to be made by the robot (‘turn right’, ‘turn left’ or ‘go straight’) and consequently this variable modifies the geometric parameters of the limit cycle. As shown in the paper, the combined use of the two techniques fills the conceptual gaps that we have identified for each of them, leading to a significant improvement in the resulting behaviour of the robot.

\section{BACKGROUND}

This section introduces the methods employed in this paper, detailing
each component individually and then combining them in the proposed
approach, which integrates potential fields with limit cycles and
nonlinear opinion dynamics.

\subsection{Nonlinear Opinion Dynamics} \label{opinion}

The nonlinear opinion dynamics model \cite{tunable_opinion_leonard} 
is a technique to emulate the process of opinion formation based on the communication of verbal and non-verbal information between humans.
In human-aware motion planning and navigation, this method can be used to enable the robot to adopt human-like navigation protocols and quickly form a decisive opinion about whether to pass an approaching human on the left or the right side \cite{opinion_leonard}, thus enabling a real-time trajectory modification to avoid collisions. This process is driven by a variable modelling the robot's attention to social cues. When the attention exceeds a threshold, the state associated with the neutral opinion is driven to a new equilibrium associated with a directional preference.

Consider a system where each agent $i$ forms opinions about two options. The opinion $z_i \in \mathbb{R}$ represents agent $i$’s preference for the first option if $z_i > 0$, and for the second one if $z_i < 0$. A neutral opinion is represented by $z_i = 0$. The continuous update of agent $i$’s opinion is described by the following equation:
\begin{equation}
\dot{z}_i = -d_i z_i + u_i \tanh \left( \alpha_i z_i + \gamma_i \sum_{k \neq i} a_{ik} z_k + b_i \right),
\end{equation}
where $d_i > 0$ determines how quickly the influence of past opinions
decays, $u_i \geq 0$ represents the agent’s attention level,
$\alpha_i > 0$ and $\gamma_i \in \mathbb{R}$ weight the influence of
the agent’s own opinion and the opinions of others, $a_{ik}$ is a
binary parameter indicating whether agent $i$ can observe agent $k$,
and $b_i$ is an internal bias or external stimulus.

This model can be extended specifically to the robot navigation framework~\cite{opinion_leonard}, focusing on forming opinions to drive its motion through a space with human movers. The evolution of the robot's opinion $z_r$ about its passing direction (left if $z_r > 0$, right if $z_r < 0$) is given by:
\begin{equation}
\dot{z}_r = -d_r z_r + u_r \tanh (\alpha_r z_r + \gamma_r \hat{z}_h + b_r),
\end{equation}
where $\hat{z}_h = \tan(\eta_h)$ approximates the human’s perceived opinion on passing direction, with $\eta_h$ being the relative heading of the human.
The robot’s attention $u_r$, critical to overcome deadlocks, is
modelled dynamically in this way:
\begin{equation}
\tau_u \dot{u}_r = -u_r + g(\kappa, \chi, R_r),
\end{equation}
where $g$ is a function proportional to human-robot distance, as $\kappa$ and $\chi$ measure the relative approach of the human and $R_r$ is a critical distance parameter.
A significant advantage of this approach is the guaranteed
deadlock-free navigation. As the robot’s attention $u_r$ surpasses the
critical value $R_r$, the neutral opinion $z_r = 0$ becomes unstable,
and the system goes through a pitchfork bifurcation, leading to one of two
stable equilibrium points (strong opinions for passing left or right). This
mechanism ensures that the robot decides on the passing direction, even in the absence of clear
cues from the human.
This proactive mechanism ensures that the robot dynamically adapts to new encounters and maintains efficient progress towards its goal.

However, in~\cite{opinion_leonard}, the robot’s heading $\theta_r$
changes according to a parameter angle $\beta_r$ adjusting the robot’s
deviation from its direct path and balancing between reliable
collision avoidance and path efficiency. This choice is not guaranteed to be collision-free  in highly
complex scenarios or for incorrect parameter settings for $\beta_r$.
Indeed, if $\beta_r$ is too low, the difference between the reference and actual trajectories will be minimal, but this will result in a shorter distance between the robot and the human, with potential risks. Conversely, if it is too high, there is
more safety, but less efficiency. For this reason, a trade-off between
efficiency and safety is required, but an optimal value for $\beta_r$
could be difficult to find, especially when human movements are
unpredictable and the lack of cooperation leads to a critical passage.
As discussed in the next sections, to overcome these limitations, we integrate opinion dynamics with potential fields and limit cycles, always ensuring collision avoidance thanks to the addition of a vortex repulsive field around humans.

\begin{figure*}[t]
\centering
    \includegraphics[width=1\textwidth]{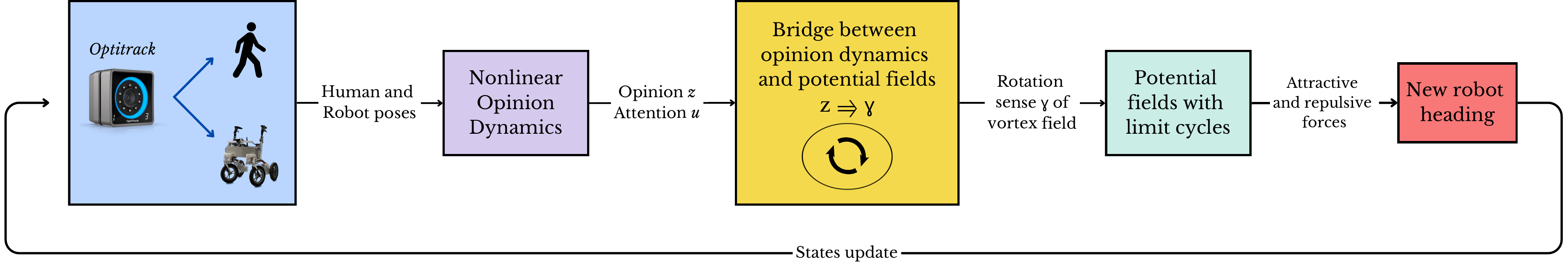}
\caption{\label{fig:diagram} The general diagram of our framework.}
\end{figure*}

\subsection{Potential Fields and Oval Limit Cycles} \label{potentials}

The combination of classical potential fields with limit cycles, described in~\cite{ovalBoldrer}, aims at facilitating the navigation in dynamic environments filled with obstacles and pedestrians. By merging these methods and introducing an innovative oval shape for the limit cycles, the paths generated are not only short and smooth, but also socially acceptable and comfortable, especially for assistive robots. This approach respects the safety and psychological comfort of humans by maintaining a clear distance from their personal space.

First, the robot moves by following the gradient direction of the potential function, which naturally attracts it to its goal, while dynamic obstacles are treated in this context as sources of repulsive fields. However, the use of repulsive potentials alone may have some limitations, especially with moving obstacles, as the trajectories generated can be jerky and potentially leading to oscillatory behaviours. To avoid these problems, a vortex field is used to steer the robot towards a limit cycle, which creates a closed trajectory around the obstacles as the robot approaches them. 

The use of oval shapes instead of the traditional circular ones
\cite{limitCycles} for the limit cycles reduces the trajectory
curvature, minimises the jerk, and respects the humans' personal
space.
The oval limit cycle is governed by the implicit equation:
\begin{equation}
\rho = 1 - \left(\frac{x_1}{b_1}\right)^2 - \left(\frac{x_2}{b_2}\right)^2 e^{\nu x_1},
\end{equation}
where $b_1$ and $b_2$ are the semi-axis lengths and $\nu$ is a deformation parameter (the oval reduces to an ellipse when $\nu = 0$). The equations for the oval trajectory, with feedforward and feedback components, are derived as follows:
\begin{equation}
\begin{aligned}
\mathbf{x}_{\text{ff}} &= \left[ -b_1 \frac{x_2}{b_2} e^{\nu/2 x_1}; \;
b_2 \frac{x_1}{b_1} e^{-\nu/2 x_1} + x_2^2 e^{\nu/2 x_1} \frac{\nu}{2 b_1 b_2} \right], \\
\mathbf{x}_{\text{fb}} &= \left[ x_1 - x_t; x_2 \right] \rho .
\end{aligned}
\end{equation}
This leads to the state evolution equations:
\begin{equation}
\begin{aligned}
\dot{x}_1 &= -\gamma x_2 e^{\nu/2 \bar{x}_1} + \alpha_1 (x_1 - x_t) \rho, \\
\dot{x}_2 &= \gamma \left( \bar{x}_1 e^{-\nu/2 \bar{x}_1} + x_2^2 e^{\nu/2 \bar{x}_1} \frac{\nu}{2} \right) + \alpha_2 x_2 \rho,
\end{aligned}
\end{equation}
where $\alpha_1$ and $\alpha_2$ are constants, $\gamma$ determines the rotation direction, and $x_t$ is a fixed distance representing the oval's centre translation along the major axis to strategically position the limit cycle with the human in the largest part of the oval, ensuring effective obstacle avoidance. 
The term \(\bar{x}_1\) represents the translated version of \(x_1\)
along the $x$ axis to account for the relative position and
orientation of the obstacle w.r.t. the oval's centre. Hence, in the
state evolution equations, \(\bar{x}_1\) is used to adjust the
feedforward dynamics so that the trajectories are centred around the
translated oval.
The term \( x_{\text{ff}} \) ensures that the solutions spiral into oval orbits with a counterclockwise rotation around the obstacle. The term \( x_{\text{fb}} \) provides feedback to pull the solutions towards the limit cycle: it acts repulsively when inside the oval (\( \rho > 0 \)), and attractively when outside the oval (\( \rho < 0 \)).

The orientation of the oval's major axis also plays a crucial role in the avoidance manoeuvre. A practical approach to setting this orientation is to align the oval's heading towards the robot, namely along the human-robot axis.
This reasoning follows from a psychological study according to which, in general, humans feel safer if the ``bypass" trajectory occurs as early as possible, respecting the proxemics limit. In this sense, regardless of the human's direction of advance, the robot starts to deviate as soon as possible. 
%

\section{OUR METHODOLOGY}\label{MyApproach}

\begin{figure}[t]
        \centering
        \subfigure[ $z < 0$, $\gamma = 1$]{\includegraphics[width=0.48\linewidth]{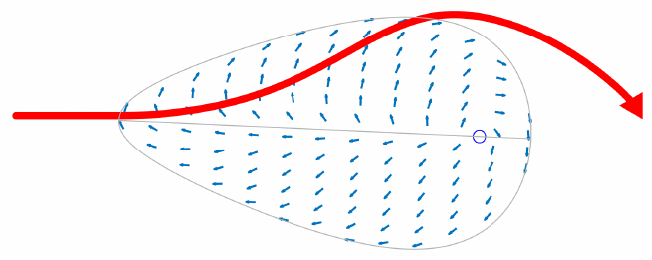}}
        \subfigure[$z > 0$, $\gamma = -1$]{\includegraphics[width=0.48\linewidth]{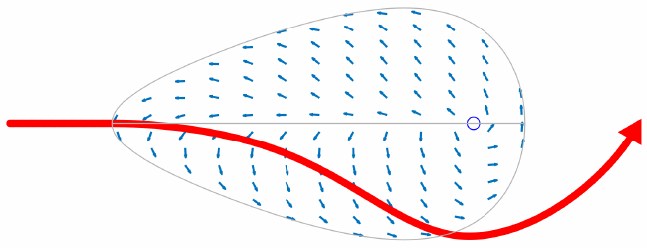}}
        \caption{Passage preference, i.e., rotation direction $\gamma$, depending on the value of $z$: if $z < 0$, then $\gamma = 1$, i.e. the vortex field is clockwise; vice versa, if $z > 0$, then $\gamma = -1$, i.e. the vortex field is counterclockwise.}  \label{fig:vortex}
\end{figure}

\begin{algorithm}
\caption{Robot's Heading Angle $\theta$ Computation}
\label{pseudocode}
\label{alg:heading_angle}
\SetAlgoLined

\textbf{Initialise }{robot's attention and opinion $z$, $u$}

\textbf{Set }{parameters for oval shape and opinion dynamics}

\textbf{Set }{initial robot and human states: $x_r$, $x_h$, $\theta$, $\theta_h$}

\textbf{Set }{goal positions: $x_{rg}$, $x_{hg}$}

\While{robot hasn't reached the goal}{
    Calculate relative heading of the human $\eta_h$\;
    
    \eIf{$\cos(\eta_h) \leq 0.5$ \tcp*[h]{Obstacle outside the robot's field of view}}{
        Reset $z$ and $u$ to their neutral values\;
    }{
        Update $\eta_h$\;
    }
    
    Update robot's opinion and attention $z$, $u$\;
    
    \eIf{$z \geq 0$}{
        $\gamma \leftarrow -1$ \tcp*[h]{CCW rotation}
    }{
        $\gamma \leftarrow 1$ \tcp*[h]{CW rotation}
    }
    
    Calculate attractive force $F_{\text{att}}$ toward goal\;
    Calculate repulsive force $F_{\text{rep}}$ from the human\;
    
    Define the oval limit cycle $\rho$\;
    \eIf{$\rho > 0$ \tcp*[h]{Inside the obstacle}}{
        Compute feedback term $x_{\text{fb}}$\;
        Compute feedforward term $x_{\text{ff}}$\;
    }{
        $x_{\text{fb}} \leftarrow [0; 0]$\; $x_{\text{ff}} \leftarrow [0; 0]$\;
    }
    
    $F_{\text{rep}} \leftarrow \gamma \cdot x_{\text{ff}} + \alpha \cdot x_{\text{fb}}$\;
    
    Calculate resultant angular velocity:
    $\omega \leftarrow \arctan2(F_{y_{\text{att}}} + F_{y_{\text{rep}}},\ F_{x_{\text{att}}} + F_{x_{\text{rep}}}) - \theta$\;
    
    Update robot's heading angle $\theta$\;
    Update robot's and human's position $x_r$, $x_h$\;
}
\end{algorithm}

As stated in Section~\ref{opinion}, the pure opinion dynamics method does not always generate trajectories respecting human comfort and can lead to collisions if the parameters are not properly configured or if the scenarios are particularly complex. The potential field strategy, on the other hand, does not allow to choose the rotation direction around the vortex field, sometimes resulting in paths that are not convenient or too long.
To address these limitations, we enhance the potential fields and the limit cycles around obstacles, which are ovally-shaped for the benefits detailed in Section~\ref{potentials}, and we incorporate the concepts of opinion and attention to further refine the navigation strategy. By integrating these concepts, we can dynamically adjust the passing direction and the repulsion intensity based on human interactions. This integration aims to create a more adaptive and responsive system, improving safety and efficiency in navigation through complex environments.

The idea is to use the opinion concept described in Section~\ref{opinion} to determine the direction of rotation around the human and, thus, the direction of the vortex field as conveniently and comfortably as possible. In particular, the opinion value \( z \) governs the passage preference, which, depending on whether \( z \) is greater or less than 0, will result in either a left or right passage. In the context of potential fields, this preference is translated into a choice of direction, which in turn depends on the parameter \( \gamma \) introduced in Section~\ref{potentials}. This parameter will be -1 or 1, meaning that the rotation will be counterclockwise or clockwise, respectively, if \( z \) is greater or less than 0. 
For more clarity on notation and procedure, refer to Figures~\ref{fig:diagram} and~\ref{fig:vortex}.

The level of attention \( u \) can also be used to determine the intensity of the vortex field, i.e., the repulsive coefficient \( k_{\text{rep}} \). As \( u \) is directly proportional to the distance between the human and the robot, the robot must pay more attention to the pedestrian as it comes closer. Similarly, the repulsive action increases with proximity, meaning \( k_{\text{rep}} \propto u \). In this way, the force pushing the robot's trajectory toward the limit cycle depends on the level of attention. 

The Algorithm~\ref{pseudocode} describes how our methodology extends these concepts by integrating nonlinear opinion dynamics with potential fields and oval limit cycles. Initially, the robot’s attention \( u \) and opinion \( z \) are set, along with parameters governing the geometry of the oval limit cycle. 
During navigation, 
if the human is outside the robot’s field of view, indicated by the relative heading \( \eta_h \), the robot resets its attention and opinion to neutral values. Otherwise, it updates \( \eta_h \) and adjusts its opinion and attention accordingly based on human's position.
The robot’s opinion \( z \) determines how it should rotate around the human. The attractive force \( F_{\text{att}} \) towards the goal and the vortex repulsive force \( F_{\text{rep}} \) of the robot from the human are then computed. The repulsive force is guided by the oval limit cycle \( \rho \), which shapes the robot’s trajectory around obstacles. When the robot is inside the limit cycle, feedback \( x_{\text{fb}} \) and feedforward \( x_{\text{ff}} \) terms are computed to refine the movement, while, if the robot is outside the limit cycle, these terms are minimised.
The repulsive force \( F_{\text{rep}} \) is updated dynamically based on $z$ and $u$, ensuring that its trajectory is adjusted smoothly and safely. Finally, the robot's heading angle \( \theta \) and position are updated, ensuring continuous adaptation. This process continues until the robot reaches its goal, dynamically balancing the need for safety, efficiency, and social comfort during navigation.
The algorithm can also be implemented in the multi-agent scenario. When there are multiple humans, multiple vortex fields have to be applied accordingly. Concerning opinion, the approach is the same as before, but at any given time, the robot considers only the closest nearby human.

In summary, our proposed approach combines the robustness of potential fields and oval limit cycles with the adaptability of opinion dynamics, offering a novel solution to enhance the safety and efficiency of autonomous navigation in complex environments.

\begin{figure}[t]
\begin{center}
        \centering
        \subfigure[]{\includegraphics[width=0.495\linewidth]{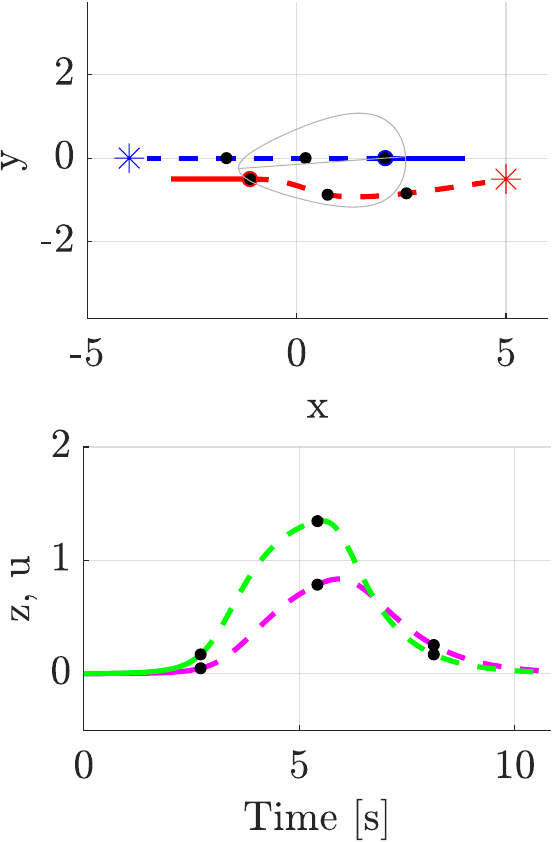}}
        \subfigure[]{\includegraphics[width=0.49\linewidth]{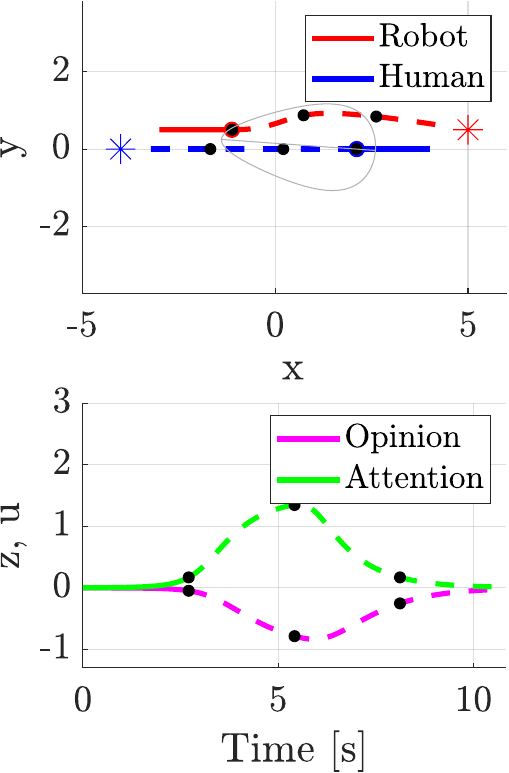}}
        \caption{ Human-robot passing simulation: human and robot trajectories towards their targets (stars). The lines change from solid to dashed when the robot enters the limit cycle, i.e. when its attention level starts to rise.
 }\label{fig:3opinions}
\end{center}
\end{figure}

\section{EXPERIMENTS AND RESULTS}

\begin{figure*}[t]
\begin{center} 
\centering
\subfigure[Nonlinear opinion dynamics]{\includegraphics[width=0.32\linewidth]{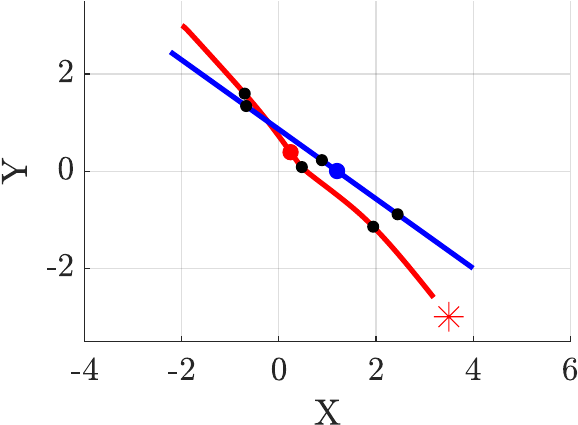}}
\subfigure[Potential fields with oval limit cycles]{\includegraphics[width=0.32\linewidth]{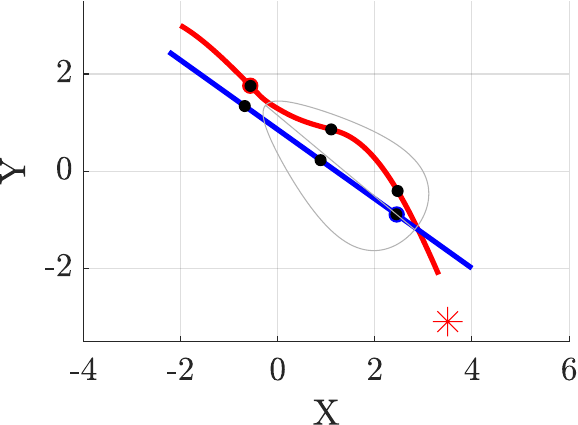}}
\subfigure[Combination of the two methods]{\includegraphics[width=0.32\linewidth]{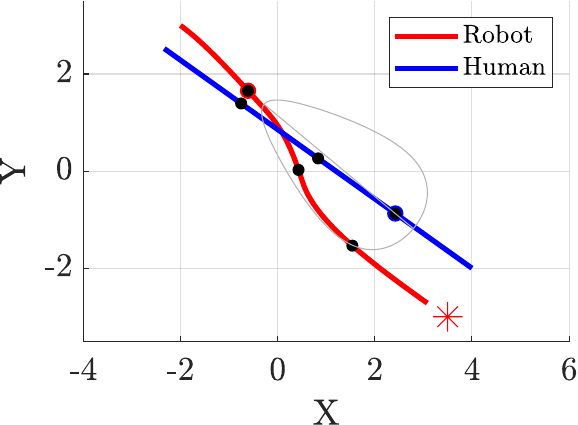}}
\caption{Comparison between approaches: red line for robot trajectory, blue line for human trajectory, red star for robot target. }\label{fig:comparison}
\end{center}
\end{figure*}

In this section, we analyse and interpret the results of the proposed navigation technique, supported by both qualitative and quantitative data from simulations and experiments. 
%
%
The parameters used for both cases are the following: regarding the form of the oval shape, \( b_1 = 2.5 \), \( \epsilon = \frac{b_1}{b_2} = 0.5 \), \( \nu = 0.5 \), \( \alpha_1 = 0.5 \), \( \alpha_2 = 5 \), while regarding the parameters controlling the opinion dynamics, \( d_r = 1.5 \), \( \alpha_r = \gamma_r = 100 \), \( \underline{u} = 0 \), \( \bar{u} = 1.5 \), \( R_r = 3 \), and \( n = 7 \). At any given moment, the robot only takes into account the closest human observed within the oval limit cycle and within an angular range of \( \left( -\frac{\pi}{3}, \frac{\pi}{3} \right) \) relative to its heading.

\begin{table}[t]
\caption{Comparison of validation metrics on 10 tests}
\centering
\resizebox{\columnwidth}{!}{%
\begin{tabular}{| c || c | c |}
\hline
& Percent increase of the & Average minimum distance \\
& robot’s path length & between the robot and human \\ 
\hline \hline
{Opinion dynamics} & 2.9294 & 0.4408 \\ \hline
{Potential fields} & 5.2035 & 0.8283 \\ \hline
{Proposed approach} & 4.2949 & 0.6117 \\ \hline
\end{tabular}%
}
\label{tab:metriche}
\end{table}

\subsection{Simulation results}

We begin by examining some simple scenarios to illustrate how the sense of rotation is influenced by the opinion dynamics and to highlight the adaptive nature of our approach. Specifically, we conduct simulations for the three following scenarios, two of which are shown in Figure~\ref{fig:3opinions}:

\begin{itemize}
\item The robot forms the opinion that the human is going right (or rather, to his left): in this case, the opinion value $z$ is positive, resulting in a counterclockwise rotation ($\gamma = -1$). The robot dynamically adjusts its path to navigate around the human from the right (see Figure~\ref{fig:3opinions}a).

\item The robot moves directly towards the human: if the human is directly in front of it, the robot defaults to a counterclockwise rotation ($\gamma = -1$) in order to effectively avoid the obstacle. In this scenario, the robot has to form a strong opinion about the direction to take to ensure a decisive and safe manoeuvre around the human, even if his position does not allow a precise preference to be formed.

\item The robot forms the opinion that the human is going left (or rather, to his right): here, the opinion value $z$ is negative, leading to a clockwise rotation ($\gamma = 1$). The robot adapts by moving around the human from the left (see Figure~\ref{fig:3opinions}b).
\end{itemize}
In each simulation, we observe how the robot's path is influenced by the opinion value $z$, which dictates the passing preference. The results clearly show that the robot can adapt its navigation strategy 
even if the opinion before getting very close to the human is zero.

\begin{figure}[t]
\centering
    \includegraphics[width=0.4\textwidth]{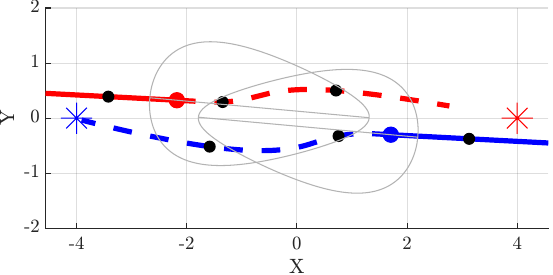}
    \caption{\label{fig:bothOvals} Simulation test with oval limit cycle for both agents. The lines change from solid to dashed when they enter the other's limit cycle.}
\end{figure}

\begin{figure*}[t]
\begin{center} 
  \centering
  \subfigure[]{\includegraphics[width=0.14\linewidth]{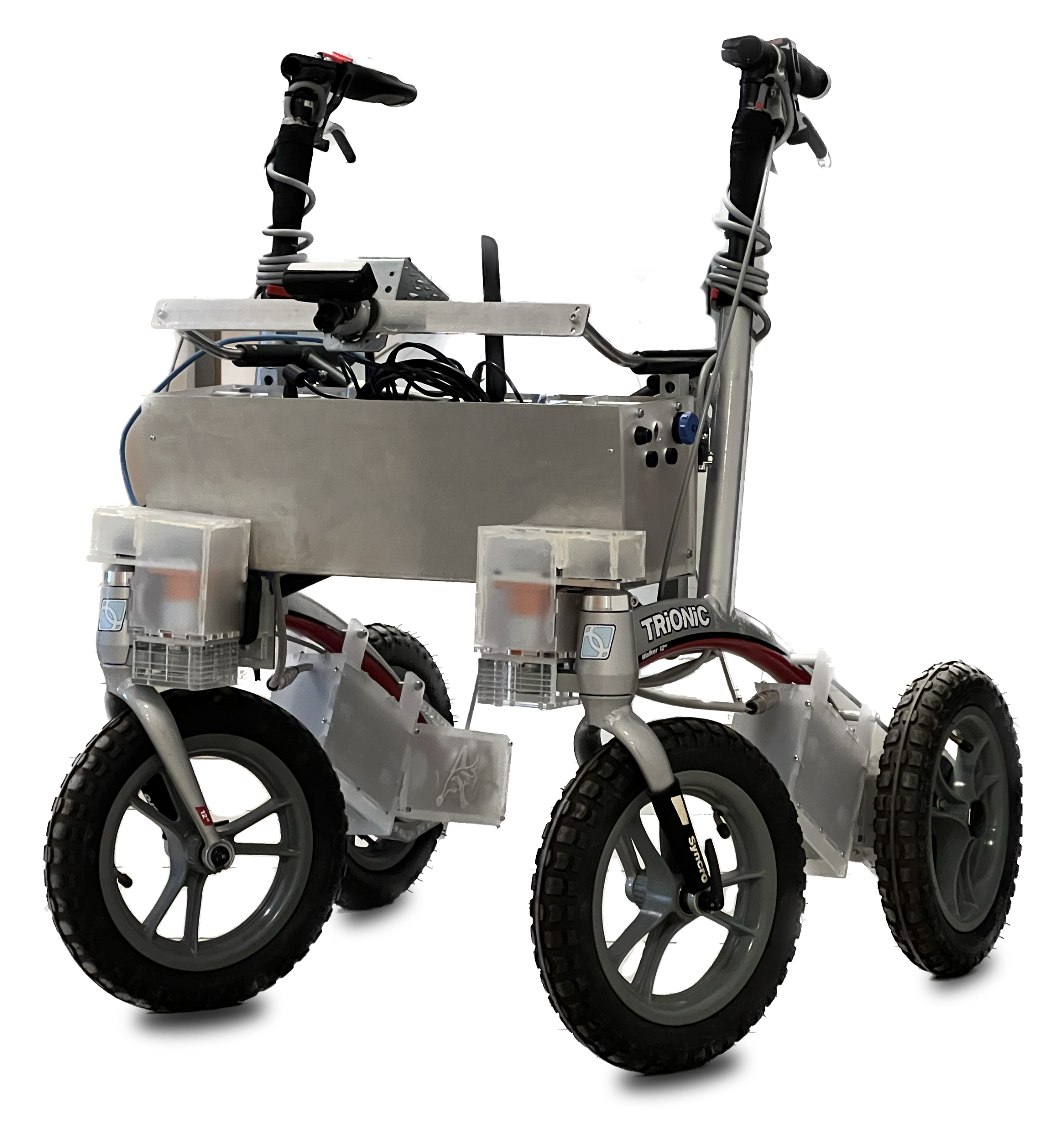}}
  \subfigure[]{\includegraphics[width=0.28\linewidth]{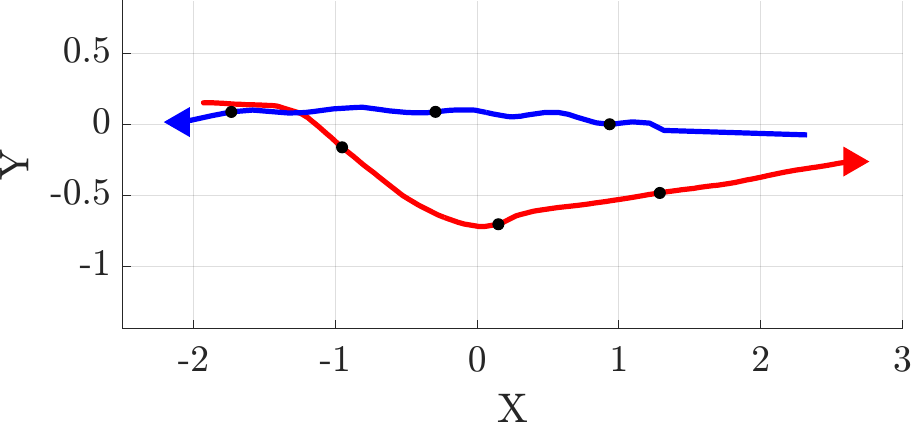}}
  \subfigure[]{\includegraphics[width=0.28\linewidth]{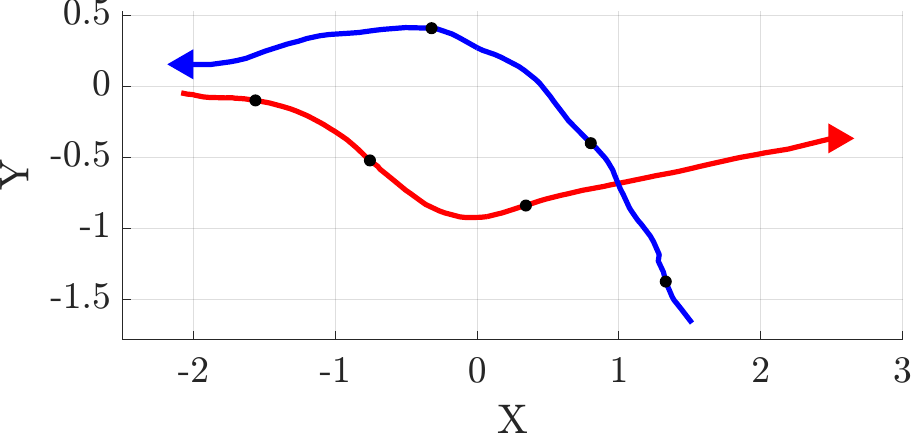}}
  \subfigure[]{\includegraphics[width=0.28\linewidth]{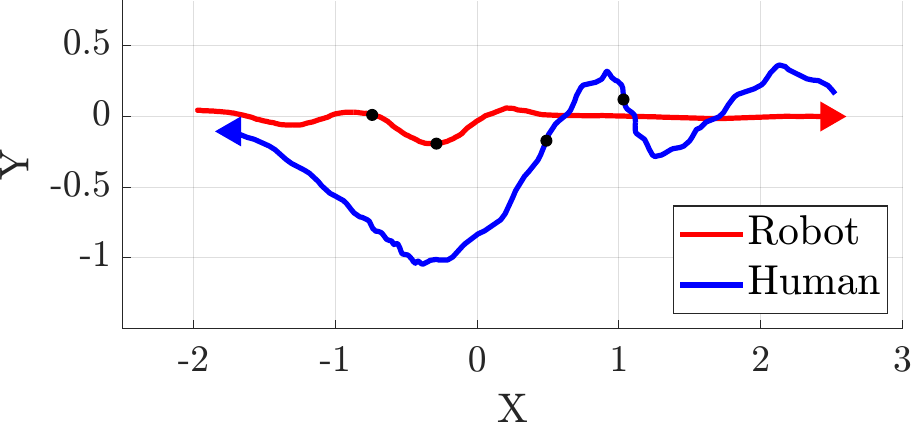}} \caption{The
    \textit{FriWalk} assistive robot used in our experiments
    (a). Trajectories obtained from three real-time experiments (b-d).}\label{fig:experiments}
\end{center}
\end{figure*}

In order to compare our approach with those proposed in~\cite{ovalBoldrer} and~\cite{opinion_leonard} and demonstrate its convenience, all three were tested in a scenario with the same parameters and agents' states. From Figure~\ref{fig:comparison}, it can be seen that the method based purely on nonlinear opinion dynamics generates a trajectory that is efficient in terms of path length, but that does not respect the psychological comfort of the human, as the overrun occurs very close to him/her. The approach based on potential fields and limit cycles instead generates a trajectory that is longer than necessary in this specific case, as the robot's target is to its right. Finally, the combination of the two methods allows the human to be overcome at an acceptable distance and in a smooth manner, even choosing to circumnavigate it in the most reasonable way. 
The goodness of the trajectories generated by our algorithm is
evaluated in terms of percentage increase of the robot's path length
and average minimum distance between robot and
human. Table~\ref{tab:metriche} shows the average results obtained by
testing the three approaches in 10 different randomly generated
scenarios, but always potentially leading to collisions.
This comparison also shows that the absence of limit cycles often generates efficient but uncomfortable trajectories, as it does not leave enough private space for the human. The use of limit cycles alone without regard to the opinion takes into account the safety distance, but often generates excessively long paths. Merging the two approaches finally allows a compromise between reliability and efficiency.

In the scenarios described above, the vortex field is only applied
around the human, who psychologically and socially needs her/his
private space to be respected by the robot. But consider, for example,
a situation where there are two humans guided by assistive robots. In
this case, there is a human presence on both sides, and the safety of
both individuals depends on their respective assistive
robots. Therefore, it makes sense to apply a limit cycle around each
agent.  In this scenario, the opinion value still remain crucial for
adjusting the passing strategy as the agents approach each other.
Indeed, when the two collide, one forms an opinion $z_1$ about the
other and the latter's opinion $z_2$ is influenced by that of the
former. This means that the opinions of the two follow the same
course. The reason for this choice is that in a collision the vortex
fields must have the same rotation direction in order to avoid each
other, otherwise they would collide. In other words, if $z_1 > 0$,
then $z_2$ also has to be greater than $0$, so that the value $\gamma$
relative to both limit cycles is $1$.
Additionally, the safety distance is dynamically regulated according to the attention level. The simulation shown in Figure~\ref{fig:bothOvals} demonstrates the enhanced capability of our method to maintain safe and efficient navigation when two robots are involved.

\subsection{Experimental setup and results}

The experimental validation of the proposed approach was carried out
using the {\em FriWalk} shown in Figure \ref{fig:experiments}a in an
indoor laboratory setting at the University of Trento. The vehicle is
equipped with electric DC motors that manage the steering angle of the
front wheels. The robot was tracked using an {\em OptiTrack} system
comprising 8 cameras, covering an area of approximately
$5 m \times 5 m$.  We then positioned retroreflective markers on both
the robot and the human to allow the {\em OptiTrack} system to
localise them \cite{naturalpoint2017}.  A ROS interface facilitated
communication, enabling the transmission of control signals to the
robot actuators and the reception of localisation data.
For each experiment, multiple trials of the same mission were performed, with variations in the general behaviour of the human experimenter.


The tests performed not only replicate the simulation conditions, but also introduce additional challenges to evaluate robot's adaptability.
In all cases, a fixed target position for the robot was set at the point (2,0), providing a consistent reference for its decision making process.
Figure~\ref{fig:experiments}b illustrates an example of a scenario where a human and a robot approach each other. The human's behaviour can vary between cooperative and unaware of the robot's presence. Regardless of the human's actions, the robot consistently demonstrates the ability to form opinions in real time and navigate efficiently, ensuring smooth interactions. The results closely match those from the simulations, confirming the effectiveness of the proposed algorithm.

To further challenge the algorithm, two additional experiments introduce erratic human behaviour.
Figure~\ref{fig:experiments}c shows a human crossing the robot's path transversely, potentially obstructing its desired path to the goal. By anticipating the human's movement and considering the oval shape of the limit cycle, the robot decides to turn to pass behind the human, thus avoiding any interference. This demonstrates the robot's ability to anticipate conflicts and adjust its path intelligently.
Figure~\ref{fig:experiments}d explores a scenario where the robot encounters an indecisive human who frequently changes direction. Initially, the robot makes small adjustments, mirroring the human's unpredictability. As they get closer, the robot's attention level increases and it becomes more decisive, reflecting its increased focus on avoiding a collision. Once past the human, the robot quickly realigns itself to its original path.
These experiments validate the robustness of the algorithm in real-world conditions, confirming the reliability of the simulation results and demonstrating the robot's potential for safe, efficient operation in environments with human interaction.

\section{CONCLUSION AND FUTURE WORK}

In this paper we present a robust framework for socially-aware robot navigation. By combining nonlinear opinion dynamics with advanced potential fields and limit cycle techniques, we achieve reliable, efficient and socially acceptable navigation and overcome the limitations of the methods taken individually.
Our methods ensure that robots can move proactively adapting to human presence, avoiding collisions due to the vortex field in the limit cycle. The use of opinion dynamics makes it possible to avoid deadlock situations by always circumventing the obstacle in the most convenient way possible.
In the future, we intend to deepen our approach in order to recognise human groups, improve adaptability and sensitivity to changes in the context and handle more complex scenarios involving more agents. Another key objective is for the robot not only to adapt as it navigates through shared spaces, but also to signal its intentions effectively. This would allow humans to form opinions about robots' behaviour and adjust their actions accordingly.

\bibliographystyle{IEEEtran}
\bibliography{root}

\end{document}